# Multi-focus image fusion using VOL and EOL in DCT domain

Mostafa Amin-Naji
Faculty of Electrical and Computer Engineering
Babol Noshirvani University of Technology
Babol, Iran
Mostafa.Amin.Naji@gmail.com

Ali Aghagolzadeh
Faculty of Electrical and Computer Engineering
Babol Noshirvani University of Technology
Babol, Iran
Aghagol@nit.ac.ir

*Abstract*—The purpose of multi-focus image fusion is gathering the essential information and the focused parts from the input multi-focus images into a single image. These multi-focused images are captured with different depths of focus of cameras. Multi-focus image fusion is very time-saving and appropriate in discrete cosine transform (DCT) domain, especially when JPEG images are used in visual sensor networks (VSN). The previous works in DCT domain have some errors in selection of the suitable divided blocks according to their criterion for measurement of the block contrast. In this paper, we used variance of Laplacian (VOL) and energy of Laplacian (EOL) as criterion to measure the contrast of image. Also in this paper, the EOL and VOL calculations directly in DCT domain are prepared using vector processing. We developed four matrices which calculate the Laplacian of block easily in DCT domain. Our works greatly reduce error due to unsuitable block selection. The results of the proposed algorithms are compared with the previous algorithms in order to demonstrate the superiority of the output image quality in the proposed methods. The several JPEG multi-focus images are used in experiments and their fused image by our proposed methods and the other algorithms are compared with different measurement criteria.

## I. INTRODUCTION

Image fusion is used to gather special and necessary information from multiple images into fewer images or preferably a single image. The fused image includes all important information from the input images, ideally, and it's more accurate explanation of the scene than each input images.

Due to limited depth of focus in cameras, it is difficult to capture an image that all components be obvious in it. This issue occurs in optical lenses of CCD/CMOS cameras [1]. Therefore, some parts of the captured images with camera sensors in visual sensor network (VSN) are blurred. In VSN there is capability to record images with different depth of focuses using several cameras [2]. The camera generate large amount of data compared to the other sensors such as pressure and temperature sensors. Also there are some limitations such as limited band width, energy consumption and processing time, which leads us to process the local input images for decreasing the amount of transmission data [3]. Therefore many researchers are seeking efficient methods for multi-focus image fusion.

Several multi-focus image fusion researches have been done in the spatial domain [4-14]. The simplest methods in the spatial domain used a weighted arithmetic mean of the source image pixels intensity [14]. This method is associated with an image blurring and decreasing the image contrast. Multi-scale image fusion methods are very common and convenient. The Laplacian pyramid transform [15], gradient pyramid-based transform [16] and the premier ones, discrete wavelet transform (DWT) [17] and shift-invariant wavelet transform (SIDWT) [18] are some examples of image fusion methods based on multi-scale transform. Because of computational complexity in multi-scale methods, they need more processing time and energy consumption. Also image fusion methods based on discrete wavelet transform need a large number of convolution operations. In addition, due to creating ringing phenomena in edge place of image, quality of the output image is reduced.

Because of aforementioned problems in multi-scale transform based image fusion methods, the researchers tended to multi-focus images fusion in DCT domain. The DCT based methods have simple calculation and are suitable for implementation in real-time applications when images are compressed in joint photographic experts group format (JPEG) [19-21]. Tang [22] proposed two image fusions techniques DCT+Average and DCT+Contrast. These methods have undesirable side effects on the output images like blurring and blocking artifact which reduces the quality of the output image. Haghighat et al. in [19] introduced the multi-focus image fusion method based on variance in DCT domain (DCT+Variance). This method divides the input images into 8×8 blocks and then creates a merged output image by selecting the corresponding blocks which have larger variance value as a criterion for evaluation of image contrast. In [23] Phamila introduced a method called DCT+AC-Max. This method selects the blocks which has more number of higher values AC coefficient in DCT domain. This method can't always choose the blocks properly, because the number of higher valued AC coefficients as a fusion criterion is invalid when the majority of AC coefficients are zero. So it makes a mistake in selection of the proper focused block. The DCT+SF method in DCT domain is introduced by Cao et al [24]. This method selects the block with higher value of the spatial frequency which is computed for each block. Although these methods (DCT+Variance, DCT+AC-Max and DCT+SF) have advantages over the previous methods, due to selection the unsuitable block from input images, the quality of the output image is reduced.

In order to improve the quality of the output fused image, new image fusion methods in DCT domain are presented in this paper. These efficient methods are based on energy of Laplacian (EOL) and variance of Laplacian (VOL). In multi-focus image fusion process, EOL and VOL are suitable criterions for measurement the contrast of image. The multi-focus image fusion using EOL is introduced in the spatial domain [7, 9]. But multi-focus image fusion methods based on DCT domain are more favorable and more useful for VSN and real-time applications. So in this article, we calculate the EOL and VOL in DCT domain using vector processing and they are used as a criterion for measure the contrast of the given image. The proposed methods increase the quality of the output image substantially.

This article is organized as follows: In section 2, the EOL and VOL calculations in DCT domain are introduced. The proposed methods are explained in section 3 and it is compared with the previous algorithms using different experiments in section 4. Finally conclusions are presented in section 5.

## II. PROPOSED METHOD

Two-dimensional DCT transform of N×N blocks of the image b(m, n) and its inverse DCT using vector processing are given as (1) and (2), respectively.

$$B = C.b.C^t \quad (1)$$
$$b = C^t.B.C \quad (2)$$

where $C$ and $C^t$ are orthogonal matrix consisted of the cosine coefficients and the transpose coefficients, respectively. B is the DCT coefficients for image's matrix of b. For C, we have:

$$C^{-1} = C^t \quad (3)$$

### A. EOL Calculation In DCT Domain

Energy of Laplacian (EOL) measures the image border sharpness and it's calculated by (4) in the spatial domain [9].

$$\text{EOL} = \sum_x \sum_y (b_{xx} + b_{yy})^2 \quad (4)$$

where

$$\begin{aligned}b_{xx} + b_{yy} = &-b(x-1,y-1) - 4 \times b(x-1,y) \\ &- b(x-1,y+1) - 4 \times b(x,y-1) + 20 \times b(x,y) \\ &- 4 \times b(x,y+1) - b(x+1,y-1) \\ &- 4 \times b(x+1,y) - b(x+1,y+1)\end{aligned} \quad (5)$$

$b_{xx} + b_{yy}$ can be computed with convolving mask (6) on 8×8 block. The size of mask (6) is 3×3 and the output matrix from convolution of the mask on the block is a 6×6 matrix. Also the EOL value is computed by the sum of the squares of 6×6 matrix elements.

| -1 | -4  | -1 |
|----|-----|----|
| -4 | +20 | -4 |
| -1 | -4  | -1 |

(6)

The 6×6 output matrix can be equivalent with the definition of four matrices which their multiplication on 8×8 block results in relation (5).

We are defining m, n, d and e matrices as below:

$$m = \begin{pmatrix} 1 & 0 & 1 & 0 & 0 & 0 & 0 & 0 \\ 0 & 1 & 0 & 1 & 0 & 0 & 0 & 0 \\ 0 & 0 & 1 & 0 & 1 & 0 & 0 & 0 \\ 0 & 0 & 0 & 1 & 0 & 1 & 0 & 0 \\ 0 & 0 & 0 & 0 & 1 & 0 & 1 & 0 \\ 0 & 0 & 0 & 0 & 0 & 1 & 0 & 1 \\ 0 & 0 & 0 & 0 & 0 & 0 & 0 & 0 \\ 0 & 0 & 0 & 0 & 0 & 0 & 0 & 0 \end{pmatrix}_{8\times 8}$$

$$n = \begin{pmatrix} -1 & 0 & 0 & 0 & 0 & 0 & 0 & 0 \\ -4 & -1 & 0 & 0 & 0 & 0 & 0 & 0 \\ -1 & -4 & -1 & 0 & 0 & 0 & 0 & 0 \\ 0 & -1 & -4 & -1 & 0 & 0 & 0 & 0 \\ 0 & 0 & -1 & -4 & -1 & 0 & 0 & 0 \\ 0 & 0 & 0 & -1 & -4 & -1 & 0 & 0 \\ 0 & 0 & 0 & 0 & -1 & -4 & 0 & 0 \\ 0 & 0 & 0 & 0 & 0 & -1 & 0 & 0 \end{pmatrix}_{8\times 8}$$

$$d = \begin{pmatrix} 0 & 1 & 0 & 0 & 0 & 0 & 0 & 0 \\ 0 & 0 & 1 & 0 & 0 & 0 & 0 & 0 \\ 0 & 0 & 0 & 1 & 0 & 0 & 0 & 0 \\ 0 & 0 & 0 & 0 & 1 & 0 & 0 & 0 \\ 0 & 0 & 0 & 0 & 0 & 1 & 0 & 0 \\ 0 & 0 & 0 & 0 & 0 & 0 & 1 & 0 \\ 0 & 0 & 0 & 0 & 0 & 0 & 0 & 0 \\ 0 & 0 & 0 & 0 & 0 & 0 & 0 & 0 \end{pmatrix}_{8\times 8}$$

$$e = \begin{pmatrix} -4 & 0 & 0 & 0 & 0 & 0 & 0 & 0 \\ -20 & -4 & 0 & 0 & 0 & 0 & 0 & 0 \\ -4 & -20 & -4 & 0 & 0 & 0 & 0 & 0 \\ 0 & -4 & -20 & -4 & 0 & 0 & 0 & 0 \\ 0 & 0 & -4 & -20 & -4 & 0 & 0 & 0 \\ 0 & 0 & 0 & -4 & -20 & -4 & 0 & 0 \\ 0 & 0 & 0 & 0 & -4 & -20 & 0 & 0 \\ 0 & 0 & 0 & 0 & 0 & -4 & 0 & 0 \end{pmatrix}_{8\times 8}$$

and

$$q = m.b.n \quad , \quad p = d.b.e \quad (7)$$

Energy of Laplacian (4), as the sum of entry-wise products of the elements, can be rewritten by the trace of a product by (8).

$$\text{EOL} = \sum_x \sum_y (b_{xx} + b_{yy})^2 = \sum_x \sum_y (q+p)^2$$
$$= \text{trace}((q+p)(q+p)^t) \quad (8)$$

where m, n, d and e are matrices that result in (5).

Defining:

$$P = C^t.p.C \quad , \quad Q = C^t.q.C \quad (9)$$
$$M = C^t.m.C \quad , \quad N = C^t.n.C \quad (10)$$
$$D = C^t.d.C \quad , \quad E = C^t.e.C \quad (11)$$

p and q can be defined as:

$$q = C.Q.C^t = C.M.C^t.C.B.C^t.C.N.C^t = C.M.B.N.C^t \quad (12)$$
$$p = C.P.C^t = C.D.C^t.C.B.C^t.C.E.C^t = C.D.B.E.C^t \quad (13)$$

From (8), (9), (10) and (11) we can find Q and P in DCT domain as (14):

$$Q = M.B.N \quad , \quad P = D.B.E \quad (14)$$

In other word, the Laplacian of block in DCT domain is computed as (15):

$$\text{Laplacian}_{DCT} = Q + P \quad (15)$$

When b is a matrix of the block values and B is its DCT, we have:

$$\text{trace}(b.b^t) = \text{trace}(B.B^t) \quad (16)$$

The EOL in DCT domain can be written as (17) using (8) and (16).

$$\text{EOL}_{DCT} = \text{trace}((Q+P)(Q+P)^t) =$$
$$\text{trace}((Q+P)(Q^t+P^t)) \quad (17)$$

where P and Q, according to (9), are DCT of q and p, respectively.

Finally, EOL in DCT domain for B (DCT representation of 8×8 block) is calculated by combining (14) and (17):

$$\text{EOL}_{DCT} =$$
$$\text{trace}((M.B.N + D.B.E) \times ((M.B.N)^t + (M.B.N)^t)) \quad (18)$$

*B. VOL Calculation In DCT Domain*

Variance of Laplacian (VOL) of image is calculated in spatial domain as (19):

$$\sigma^2 = \frac{1}{N^2} \sum_{k=0}^{N-1} \sum_{l=0}^{N-1} \text{Laplacian}^2(k,l) - \mu^2 \quad (19)$$

where μ is mean value of Laplacian of the block.

Haghighat et.al in [19] computed the variance of 8×8 block in DCT domain is as (20):

$$\sigma^2_{DCT} = \sum_{k=0}^{7} \sum_{l=0}^{7} \frac{d^2(k,l)}{64} - d(0,0) \quad (20)$$

where d(k, l) is the DCT representation of the block.

For calculate variance of image Laplacian directly in DCT domain we should replace d(k,l) in (20) with achieved Laplacian$_{DCT}$ in (15). Finally the variance of Laplacian (VOL) in DCT domain is obtained as (21):

$$\text{VOL}_{DCT} = \sum_{k=0}^{7} \sum_{l=0}^{7} \frac{\text{Laplacian}_{DCT}(k,l)}{64}$$
$$- \text{Laplacian}_{DCT}(0,0) \quad (21)$$

*C. Block Selection*

For a simple description of the proposed algorithm, two images A and B are considered. The proposed image fusion process could be extended for more than two images. It is assumed the input images were aligned by an image registration method before performing the image fusion process. The general structure of the proposed method for two images fusion is shown in Fig.1.

The region of the focused image has more information and high contrast. Subsequently this region has more raised and evident edges. The amount and intensity of edges in image is used as criterion to specify the image quality and contrast. Energy of Laplacian and variance of Laplacian are the appropriate measures to show the amount of edges in image. Therefore, the block of image which comes from focused area has higher EOL or VOL value than the block of the unfocused area. The proposed method divides the input images into 8×8 blocks. In next step, DCT coefficients of each block are calculated. The first input block and the second one are named as imA and imB, respectively. Then EOL or VOL value is computed for every block in DCT domain. Block with higher EOL or VOL value is represented as focused area and it is selected for the output fused image. The proposed algorithm makes a decision map X(i,j) as (22) by comparing the EOL or VOL value of the corresponding blocks.

$$X(i,j) = \begin{cases} 1 & \text{if } (\text{EOL or VOL})_{DCT}(\text{imA}) > (\text{EOL or VOL})_{DCT}(\text{imB}) \\ -1 & \text{if } (\text{EOL or VOL})_{DCT}(\text{imA}) < (\text{EOL or VOL})_{DCT}(\text{imB}) \\ 0 & \text{otherwise} \end{cases}$$
$$(22)$$

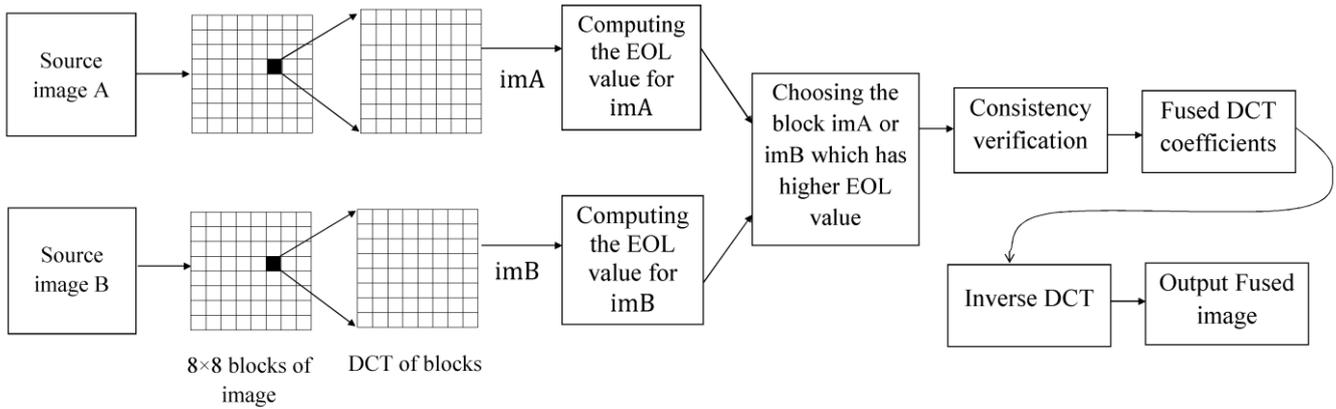

Fig. 1. General structure of proposed method.

For $X(i,j) = 1$, the block of imA is selected for output fused image. Subsequently for $X(i,j) = -1$, the block of imB is selected.

*D. Consistency Verification (CV)*

Suppose the central block of an area among selected blocks for fused image comes from image B, but the majority of neighboring blocks comes from image A. It means that the central block should belong to image A. Li et al. used the majority filter for consistency verification [17]. The central block replaced by the corresponding block from image A using a majority filter which is applied on decision map M(m,n). Therefore the consistency verification (CV) is applied as a post-processing after image fusion process to improve the quality of the output image and reduce the error due to unsuitable block selection.

## III. EXPERIMENTAL RESULTS AND ANALYSIS

This section discusses the performance of the proposed method and exhibits the results of simulation for the proposed method and also the other methods for comparison. The results of simulation for the proposed method compared with the results of the previous methods such as the methods which are based on multi-scale transform like DWT [17] and SIDWT [18], and the methods based on DCT domain like DCT+Average [21], DCT+Variance [19], DCT+ AC_Max [23] and DCT+SF [24]. The used test images in simulations shown in Fig.2 obtained from the online database [25] as the referenced images. Also the famous non-referenced multi-focus image "Disk" is used for second experiment obtained from the online database [26]. The DCT+Variance algorithm code of MATLAB simulation is taken from the online database [27]. For the wavelet based methods, DWT with DBSS (2,2) and the SIDWT for Haar basis with three levels of decomposition are considered. Simulations of these methods are done by "Image Fusion Toolbox" [26]. All of the DCT+Average, DCT+ AC_Max and DCT+SF methods are simulated by the authors in MATLAB. For the majority filter used in CV, an averaging mask of size 5×5 is considered for all experiments.

For assessment of the proposed algorithm and comparing it with the previous algorithms, evaluation performance metrics of image fusion are used. The structural similarity (SSIM) [28] and Mean-squared error (MSE) [29] are used as metrics that needs the ground truth image for the referenced images. The total information transferred from source images to the fused image ($Q^{AB/F}$), the total loss of information ($L^{AB/F}$), noise or artifacts added in fused image due to fusion process ($N^{AB/F}$) which provided by Petrovic [30], and feature mutual information (FMI) [31] are used for the non-reference images which their ground truth image is not available.

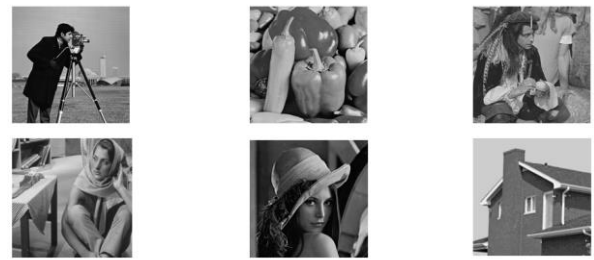

Fig. 2. Standard test images used for simulation

The first experiment conducted for applying the proposed methods and the other methods on 12 pairs of artificial multi-focus images generated from six images depicted in Fig.2. The non-focused conditions of each pairs produced by artificial blurring using two averaging masks of size 5×5 and 9×9. Blurring process is performed on right or left half of the images. The average values of SSIM and MSE for the proposed and the other algorithms are listed in TABLE I. Our proposed method shows the best results among the other methods. The second experiments conducted for assessment of the proposed algorithm and the others on real multi-focus image ("Disk") which it was captured with different depth of focuses in camera. Evaluation performance metrics $Q^{AB/F}$, $L^{AB/F}$, $N^{AB/F}$ and FMI values are showed in TABLE II. Fig.3 depicts the proposed method result image, the other methods result images and their local magnified version of "Disk" source image. Thus in the non-referenced image, the proposed method also has better results and quality.

TABLE I. THE MSE AND SSIM COMPARISION OF THE VARIOUS IMAGE FUSION APPROACHES ON REFERENCE IMAGES.

| Method | Average values for 12 pairs image created from image shown in Fig.2 | |
|---|---|---|
| | SSIM | MSE |
| DCT+Average [22] | 0.8983 | 68.9065 |
| DWT [17] | 0.9542 | 20.5945 |
| SIDWT [18] | 0.9563 | 17.3855 |
| DCT+Variance[19] | 0.9683 | 19.2826 |
| DCT+AC-Max [23] | 0.9894 | 5.1054 |
| DCT+SF [24] | 0.9870 | 6.9794 |
| **DCT+EOL (proposed)** | **0.9944** | **2.2789** |
| **DCT+VOL (proposed)** | **0.9950** | **1.8950** |
| DCT+Variance+CV [19] | 0.9895 | 10.4003 |
| DCT+AC-Max+CV [23] | 0.9965 | 1.7855 |
| DCT+SF+CV [24] | 0.9958 | 2.8608 |
| **DCT+ EOL +CV (proposed)** | **0.9972** | **1.0205** |
| **DCT+ VOL +CV (proposed)** | **0.9971** | **0.9297** |

TABLE II. THE $Q^{AB/F}$, $L^{AB/F}$, $N^{AB/F}$ AND FMI COMPARISION OF THE VARIOUS IMAGE FUSION APPROACHES ON NON-REFERENCED IMAGES.

| Method | "DISK" | | | |
|---|---|---|---|---|
| | $Q^{AB/F}$ | $L^{AB/F}$ | $N^{AB/F}$ | FMI |
| DCT+Average [22] | 0.5187 | 0.4782 | 0.0063 | 0.9013 |
| DWT [17] | 0.6302 | 0.2552 | 0.3362 | 0.9039 |
| SIDWT [18] | 0.6694 | 0.2764 | 0.1564 | 0.9049 |
| DCT+Variance[19] | 0.7165 | 0.2612 | 0.0478 | 0.9070 |
| DCT+AC-Max [23] | 0.6763 | 0.2910 | 0.0696 | 0.9057 |
| DCT+SF [24] | 0.7213 | 0.2600 | 0.0415 | 0.9086 |
| **DCT+EOL (proposed)** | **0.7271** | **0.2522** | **0.0444** | **0.9093** |
| **DCT+VOL (proposed)** | **0.7272** | **0.2521** | **0.0444** | **0.9094** |
| DCT+Variance+CV [19] | 0.7192 | 0.2734 | 0.0163 | 0.9100 |
| DCT+AC-Max+CV [23] | 0.6990 | 0.2922 | 0.0186 | 0.9102 |
| DCT+SF+CV [24] | 0.7269 | 0.2662 | 0.0153 | 0.9106 |
| **DCT+ EOL +CV (proposed)** | **0.7291** | **0.2664** | **0.0096** | **0.9109** |
| **DCT+ VOL +CV (proposed)** | **0.7288** | **0.2666** | **0.0097** | **0.9108** |

## IV. CONCLUSION

Two new multi-focus image fusion methods in DCT domain based on energy of Laplacian (EOL) and variance of Laplacian (VOL) were introduced in this paper. We calculate EOL and VOL directly in DCT domain using vector processing. Considering EOL and VOL as a measurement of image contrast in multi-focus image fusion process caused better results compared with the previous works. Also due to simple implementation of the proposed algorithms in DCT domain, it is appropriate for using in real-time applications. Accuracy of the proposed methods is assessed by applying the proposed algorithms and the others on the several referenced images and one non-referenced image by evaluating the results with the various performance metrics. The results show superiority of the output image quality for the proposed algorithms comparison to some precious algorithms.

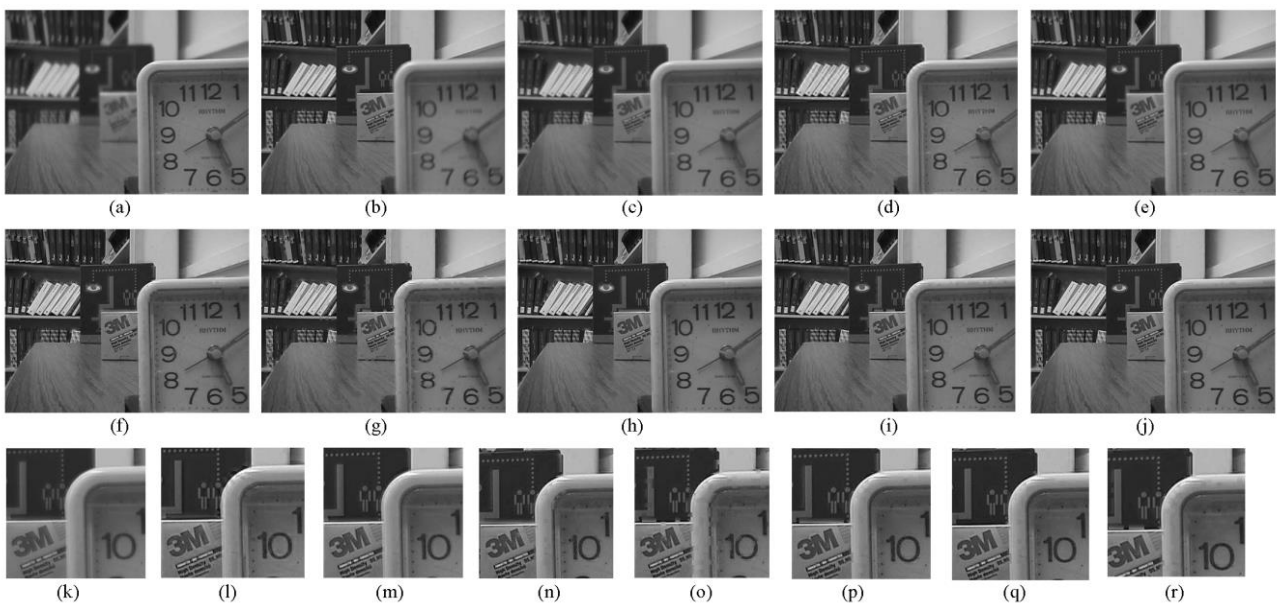

Fig. 3. Source images "DISK" and the fusion results. (a) The first source image with focus on the right. (b) The second source image with focus on the left. (c) DCT + Average result. (d) DWT result. (e) SIDWT result. (f) DCT + Variance result. (g) DCT + Ac-Max result. (h) DCT + Spatial frequency result. (i) The result of DCT + EOL (Proposed). (j) )The result of DCT + VOL (Proposed). (k), (l), (m), (n), (o), (p), (q) and (r) are the local magnified version of (c), (d), (e), (f), (g), (h), (i) and (j), respectively.